\documentclass[letterpaper, 10 pt, conference]{ieeeconf}  

\IEEEoverridecommandlockouts                              

\usepackage{graphicx}
\usepackage{algorithm}
\usepackage{algorithmic}
\usepackage{amsmath} 
\usepackage{amssymb}  
\usepackage{subcaption}
\usepackage{booktabs}
\usepackage{float}

\title{\LARGE \bf
Hierarchical Consensus-Based Multi-Agent Reinforcement Learning for Multi-Robot Cooperation Tasks

}

\author{Pu Feng$^{1}$, Junkang Liang$^{2}$, Size Wang$^{2}$, Xin Yu$^{1}$, Xin Ji$^{3}$, \\ Yiting Chen$^{3}$, Kui Zhang$^{1}$, Rongye Shi$^{1,2}$, and Wenjun Wu$^{1,2}$%
\thanks{1 State Key Laboratory of Complex \& Critical Software Environment, Beihang University. 
2 School of Artificial Intelligence, Beihang University. 
3 State Grid Corporation of China. Emails: \{fengpu, liangjunkang, sizewang, nlsdeyuxin, shirongye, wwj09315\}@buaa.edu.cn. Wenjun Wu is the corresponding author.}
}

\begin{document}

\maketitle
\thispagestyle{empty}
\pagestyle{empty}

\begin{abstract}

In multi-agent reinforcement learning (MARL), the Centralized Training with Decentralized Execution (CTDE) framework is pivotal but struggles due to a gap: global state guidance in training versus reliance on local observations in execution, lacking global signals. Inspired by human societal consensus mechanisms, we introduce the Hierarchical Consensus-based Multi-Agent Reinforcement Learning (HC-MARL) framework to address this limitation. HC-MARL employs contrastive learning to foster a global consensus among agents, enabling cooperative behavior without direct communication. This approach enables agents to form a global consensus from local observations, using it as an additional piece of information to guide collaborative actions during execution. To cater to the dynamic requirements of various tasks, consensus is divided into multiple layers, encompassing both short-term and long-term considerations. Short-term observations prompt the creation of an immediate, low-layer consensus, while long-term observations contribute to the formation of a strategic, high-layer consensus. This process is further refined through an adaptive attention mechanism that dynamically adjusts the influence of each consensus layer. This mechanism optimizes the balance between immediate reactions and strategic planning, tailoring it to the specific demands of the task at hand. Extensive experiments and real-world applications in multi-robot systems showcase our framework's superior performance, marking significant advancements over baselines.

\end{abstract}

\section{INTRODUCTION}
Multi-Agent Reinforcement Learning (MARL) is garnering increasing attention for its capability to tackle complex tasks~\cite{shi2020improving,shi2021physics1}. Tasks involving distributed multi-robots~\cite{yahya2017collective,shi2021physics} often require several agents to collaborate based on their local observations to accomplish a given objective. This requirement aligns with the commonly adopted MARL framework of Centralized Training with Decentralized Execution (CTDE), as exemplified by methods such as MADDPG (Multi-Agent Deep Deterministic Policy Gradient)~\cite{lowe2017multi} and MAPPO (Multi-Agent Proximal Policy Optimization)~\cite{yu2022surprising}. These approaches utilize global information during the training phase through the critic, while the actor relies solely on individual observations during execution. This setup presents a substantial challenge: agents fail to reach a consensus during task execution, which impedes their collective efficacy in cooperation.


In addressing this issue, three primary methodologies have been advanced. The first involves strategies for communication-based multi-agent reinforcement learning~\cite{sheng2022learning}, which are faced with challenges in selective information sharing and increased bandwidth requirements. The second method leverages intrinsic rewards to create leader-follower dynamics~\cite{wen2021optimized}, yet this approach is constrained by its task-specific effectiveness and encounters difficulties with broad applicability and generalization. The third approach, employing mean field theory~\cite{yang2018mean}, offers a promising direction but often struggles to effectively handle complex tasks. Given existing methods' limitations, sociology's research~\cite{cook2016consensus} on consensus team agents interacting to align on shared values offers potential solutions for CTDE's challenges with partial observations.


\begin{figure}[t]
    \centering
    \includegraphics[scale=0.5]{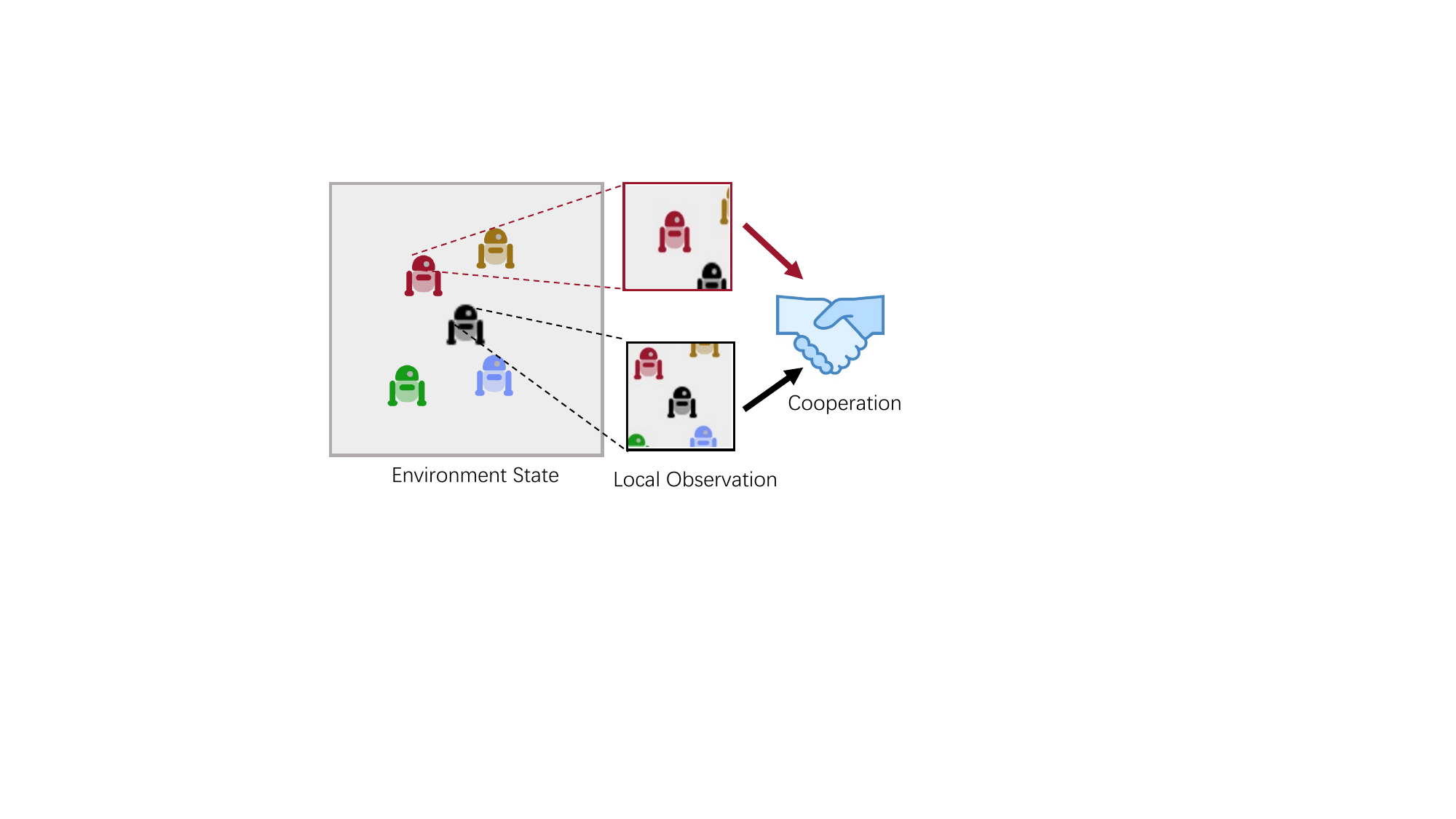}
    \caption{The relationship between the environmental state and local observations in the CTDE framework. Despite differing local observations, they all correspond to the same environmental state at each timestep, providing diverse perspectives of a unified global state. In traditional CTDE approaches, agents rely solely on these local observations for decision-making during execution.}
    \label{Intro}
    \vspace{-0.25in}
\end{figure}

Inspired by consensus mechanisms in multi-agent systems~\cite{qin2016recent}, we introduce the Hierarchical Consensus-based Multi-Agent Reinforcement Learning (HC-MARL) framework, designed to facilitate substantive multi-agent collaboration in settings characterized by local observations and the absence of direct communication. As shown in Fig.~\ref{Intro}, despite differing local observations, agents all correspond to the same environmental state at each timestep, merely offering diverse perspectives of a unified global state. Adapting ideas from contrastive learning~\cite{le2020contrastive}, we first map local observations into discrete latent spaces as forms of invariances using the consensus builder. We define these invariances as global consensus. This global consensus is then treated as an additional piece of local observation information fed into the actor network. Notably, utilizing the consensus only requires an agent's local observations, aligning with the CTDE framework's prerequisite for partial observability. 

\begin{figure}[t]
    \centering
    \includegraphics[scale=0.38]{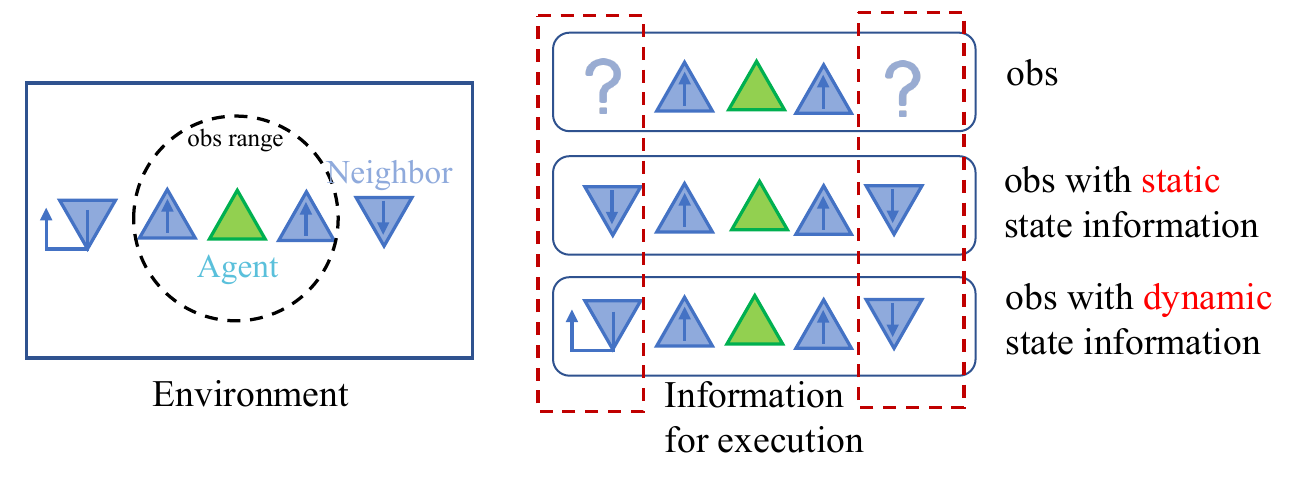}
    \caption{Importance of Dynamic State Information: The diagram illustrates agents as green triangles and neighbors as blue triangles. The orientation of agents is indicated by the vertical position of the triangles, and their motion direction is shown by the arrows. The left side displays the environmental state, while the right side shows information usable in execution within CTDE. Static environmental information provides position and orientation, whereas dynamic information additionally offers speed data.}
    \label{hc}
    \vspace{-0.2in}
\end{figure}

The initial findings, however, highlighted a limitation: relying exclusively on observations from a single timestep to establish a single-layer consensus falls short in fully capturing the nuances of sequential tasks. For example, as depicted in Fig.~\ref{hc}, within the CTDE paradigm, the agents' execution information is confined to local observations. Incorporating static environmental states, or short-term consensus, supplements execution by integrating additional static information from other agents, such as their positions and orientations, not initially available. To effectively incorporate dynamic attributes, like other agents’ velocities, introducing dynamic state information through long-term consensus becomes crucial. To address this, we propose the hierarchical consensus mechanism, where consensus based solely on single-step observations is defined as low-layer consensus, focusing on short-term optimality. High-layer consensus, on the other hand, aggregates local observations from multiple timesteps, forming a long-term group consensus that considers the broader strategic trends or states of multiple agents. Lastly, we employ an attention mechanism to dynamically weigh the influence of each consensus layer for collaborative needs, ensuring that the system can adaptively prioritize either short-term or long-term considerations based on the evolving context of the task. This hierarchical consensus serves as additional local observation during the execution of actions, providing agents with context-rich group behavioral insights.

Our HC-MARL framework can be seamlessly integrated as components of almost all MARL algorithms. The contributions of this paper are summarized as follows:
\begin{itemize}

\item
Leveraged contrastive learning to construct global consensus from local observations, enhancing cooperative action execution in multi-agent reinforcement learning.

\item
Introduced the HC-MARL framework with a hierarchical consensus mechanism, featuring both short-term and long-term consensuses. 

\item 
Implemented an adaptive attention mechanism that dynamically tunes the influence of each consensus layer, optimizing the balance between immediate responses and strategic planning according to the specific demands of the task.

\item
Demonstrated the superior performance of our framework over baselines through extensive evaluations in both simulated tasks and real-world robot experiments.
\end{itemize}


\section{Related Work}

\subsection{Consensus in Multi-Agent System}
Consensus means the interaction between groups of agents in a team to reach an agreement on a common value or state~\cite{amirkhani2022consensus}. Consensus in Multi-Agent Systems has garnered extensive research interest, primarily focusing on achieving shared agreement among agents. Research in this area spans three main categories: first, studies inspired by biological mechanisms and wildlife collective behaviors~\cite{olfati2006flocking,li2023mir2}; second, theoretical explorations using models like the graph theory for foundational insights into consensus~\cite{ren2005consensus,peng2024graphrare}; and third, practical applications, including the development of consensus models and protocols~\cite{yu2021swarm}, with a keen focus on convergence, equilibrium, and implementation challenges. Notably, Multi-Agent Reinforcement Learning (MARL)~\cite{yu2024leveraging,yu2023esp,li2023byzantine} has become a focal point within the third category, emphasizing the utilization of consensus for enhancing agent cooperation in complex environments.

\subsection{Contrastive Learning} 


Recent studies have increasingly leveraged self-supervised learning~\cite{jaiswal2020survey} to enhance model capabilities for downstream tasks. Among these, contrastive learning~\cite{khosla2020supervised} stands out as a particularly tractable approach. It operates on the principle of minimizing the distance between augmented versions of the same sample while maximizing the distance between distinct ones, as noted by Wang~\cite{wang2020understanding}. This method effectively boosts the model's understanding of equivalent data representations. However, the collection of positive and negative samples in contrastive learning presents challenges that significantly impact its effectiveness. In reinforcement learning (RL), samples are collected through ongoing interactions with the environment, highlighting the potential benefits of integrating contrastive learning with RL. Recent research efforts~\cite{laskin2020curl}, have successfully employed contrastive learning for representation learning prior to reinforcement learning, achieving unparalleled data efficiency in pixel-based RL tasks. Moreover, contrastive learning has been utilized to formulate reward functions within RL systems~\cite{dwibedi2018learning}.

\subsection{Contrastive Learning for MARL} 
Despite the progress in single-agent RL contexts, the exploration of contrastive learning in MARL remains comparatively underdeveloped. Lin~\cite{chen2022multiagent} proposed incorporating contrastive learning outcomes as a loss function in the MARL training process, applying it to multi-agent path planning tasks. Xu~\cite{xu2023consensus} introduced using contrastive learning to articulate the observational differences among agents. Further, Liu~\cite{liu2021social} promoted the development of a common language by maximizing the mutual information between the messages of given trajectories in a contrastive manner. Our HC-MARL method employs contrastive learning to establish a multi-layer global consensus. To the best of our knowledge, this is the first study utilizing contrastive learning to create a structured global consensus for collaborative agent behavior under local observations.

\section{Preliminaries}


\subsection{Problem Formulation}
The multi-robot cooperation task can be formulated as a decentralized partially observable Markov decision process (Dec-POMDP), defined as $(I, S, A, T, R, O, Z, \gamma)$. The index set $(I=\{1,..,N\})$ represents the set of agents. $S$ is the global state space. Note that each agent is only capable of partial observation of the environment $s \in S$, and the individual observation $o_i \in O$ comes from the local observation function $o_{i}=\mathcal{Z}_i\left(s\right): S \rightarrow O$. Each agent $i$ chooses its action according to its policy $a_{i} \sim \pi_i\left(\cdot \mid o_i\right)$. The joint action space $A$ consists of the union of all agent's action space $\bigcup_{i=1}^{N} \mathcal{A}_{i}$. We define the state transition function $T: S \times A \rightarrow S$ and the discount factor $\gamma \in(0,1)$. All agents share the same joint reward function $R(s, a)$. The agents aim to maximize the expected joint return, defined as $\mathbb{E}_\pi\left[\sum_{t=0}^{\infty} \gamma^t R\left(s_t, a_t\right)\right]$

\subsection{Centralized Training with Decentralized Execution (CTDE)}
To address the problems under decentralized partially observable Markov decision processes (Dec-POMDPs), the Centralized Training with Decentralized Execution (CTDE) framework emerges as a critical approach. CTDE delineates a methodology where the training phase is centralized, allowing agents to access global information and learn coordinated strategies. In contrast, during execution, each agent operates independently based on its local observations, aligning with the decentralized nature of many real-world applications.


Although value decomposition methods and policy-based multi-agent methods differ significantly in structure, they both adhere to the CTDE principles and encounter challenges in providing unified guidance during decentralized execution in fully cooperative tasks. To illustrate the optimization process within the CTDE paradigm, we focus on the most commonly used method, MAPPO. MAPPO is an actor-critic method based on the CTDE paradigm. Each agent learns a policy \(\pi\) in an on-policy manner. MAPPO consists of a centralized critic and several independent actors corresponding to each agent. During the centralized training phase, the critic utilizes global state information to estimate the joint action-value \(Q\). The critic is trained by minimizing the Temporal Difference (TD) error as follows:

\begin{equation}
L^{\text{Critic}}(\phi) = \mathbb{E}_{(s, \mathbf{a}, r, s')} \left[ \left( Q_\phi(s, \mathbf{a}) - (r + \gamma Q_{\phi'}(s', \mathbf{a}')) \right)^2 \right]
\end{equation}

where \(Q_\phi(s, \mathbf{a})\) represents the critic's current estimate of the joint action-value for the global state \(s\) and actions \(\mathbf{a}\), parameterized by \(\phi\); \(r\) is the immediate reward received after taking action \(\mathbf{a}\) in state \(s\); \(s'\) is the next state, and \(\mathbf{a}'\) is the action taken in the next state as per the current policy; \(\gamma\) is the discount factor; and \(\phi'\) refers to the parameters of the target critic used for bootstrapping.

For the actors, which operate based on their local observations during decentralized execution, the policy gradient is adjusted to reflect their dependence on local observations \(o\). The policy gradient for optimizing each agent's policy \(\pi\), considering local observations, is derived as follows:

\begin{equation}
\nabla_{\theta} J(\pi) = \mathbb{E}_{o, \mathbf{a} \sim \rho^\pi} \left[ \nabla_{\theta} \log \pi(o, \mathbf{a}|\theta) Q_\phi(s, \mathbf{a}) \right]
\end{equation}

\subsection{Contrastive Learning}

The Knowledge Distillation with No Labels (DINO)~\cite{caron2021emerging} method offers a solution as a form of self-supervised contrastive learning that leverages a teacher-student network architecture. In this framework, for a given sample \(u\), a new sample \(u'\) is generated through data augmentation. Both \(u\) and \(u'\) are then fed into the student and teacher networks, respectively, producing classification distributions \(P_S(u)\) and \(P_T(u')\). In the absence of true labels, the teacher network's output serves as pseudo-labels, and the student network aims to optimize by minimizing the cross-entropy loss between its output and these pseudo-labels.The cross-entropy loss used for optimization can be formalized as:

\begin{equation}
L_{CL} = -\sum_{c} P_T(u')_c \log P_S(u)_c
\end{equation}

where \(c\) indexes over the classes, \(P_T(u')_c\) is the pseudo-label probability for class \(c\) produced by the teacher network for augmented sample \(u'\), and \(P_S(u)_c\) is the probability produced by the student network for the original sample \(u\).

The teacher and student networks share the same architecture, with the teacher's parameters being an exponential moving average (EMA) of the student's parameters. This arrangement facilitates a continuous refinement of the student network's learning through guidance from a slowly evolving version of itself, represented by the teacher network.

In multi-agent reinforcement learning scenarios, local observations made by different agents can be considered as diverse augmented samples of the same global state. The global consensus, therefore, corresponds to the classification output from the teacher-student network framework. By constructing this consensus metric, our aim is to guide agents, operating under local observations, towards forming global cooperation.

\section{METHODS}
In this section, we introduce Hierarchical Consensus-based Multi-Agent Reinforcement Learning (HC-MARL), a novel framework that dynamically guides agents towards cooperative execution under partial observations through a hierarchical consensus mechanism.

\subsection{Consensus Builder}

In multi-agent reinforcement learning, agents execute actions based on local observations, leading to a lack of global information guidance during execution. Inspired by Xu~\cite{xu2023consensus}, we utilized insights from his method to develop a consensus mechanism based on local observations. This section focuses on building an effective consensus within the CTDE framework. As discussed in a previous section, the DINO framework processes a sample and its data-augmented equivalent through a teacher-student network. We treat an agent's observation $o$ as the sample, and observations from other agents as the equivalent samples. In other words, the observation function $Z$ is considered an augmentation operation, where the global observation \(s\) is augmented into each agent's observation \(o\). These observations \(o\) correspond to the same global state \(s\). 

Drawing inspiration from human patterns of situational awareness---where individuals often derive a broad understanding of their environment from local cues, such as discerning general cardinal directions in a city without knowing precise coordinates---we propose a model that leverages discrete categories for consensus in multi-agent systems. This approach simulates how agents might infer a macro classification of the current state from limited, local information. Specifically, we define consensus in terms of \(K\) distinct classes, enabling the consensus module to categorize an agent's local observations into a unified class \(k\), which subsequently acts as the global consensus for guiding the agents' actions.

For each agent \(i\) in a set of \(n\) agents, the classification distribution resulting from local observations is denoted by \(P_S(o_i)\) for the student network and \(P_T(o_i)\) for the teacher network. The consensus among agents is evaluated by pairwise comparison, optimizing the Consensus Builder through minimizing the sum of cross-entropy loss:

\begin{equation}
L_{S}(\theta) = -\sum_{i=1}^{n}\sum_{j=1}^{n}\sum_{k} P_T(o_j)_k \log P_S(o_i)_k
\end{equation}

where \(i\) and \(j\) are the indices of the agents. \(P_T(o_j)_k\) and \(P_S(o_i)_k\) denote the probability of category \(k\) in the distribution output by the teacher and student networks, respectively. The consensus \(c\) for each agent is obtained as follows:

\begin{equation}
c_i = \arg\max_k P_S(o_i)_ck
\end{equation}

This procedure identifies the consensus class \(c_i\) for agent \(i\) by selecting the class \(k\) with the maximum probability in the classification distribution \(P_S(o_i)\), as generated by the student network for observation \(o_i\).

Such a method underscores the agents' collaborative push towards a harmonized environmental perception, thereby enabling a collective consensus derived from individual observations. By employing a cross-entropy loss function, the framework promotes similarity in probability distributions for observations by different agents, even in disparate local contexts, thereby fostering a unified understanding of the global state.

\begin{figure}[t]
\centering
\includegraphics[scale=0.35]{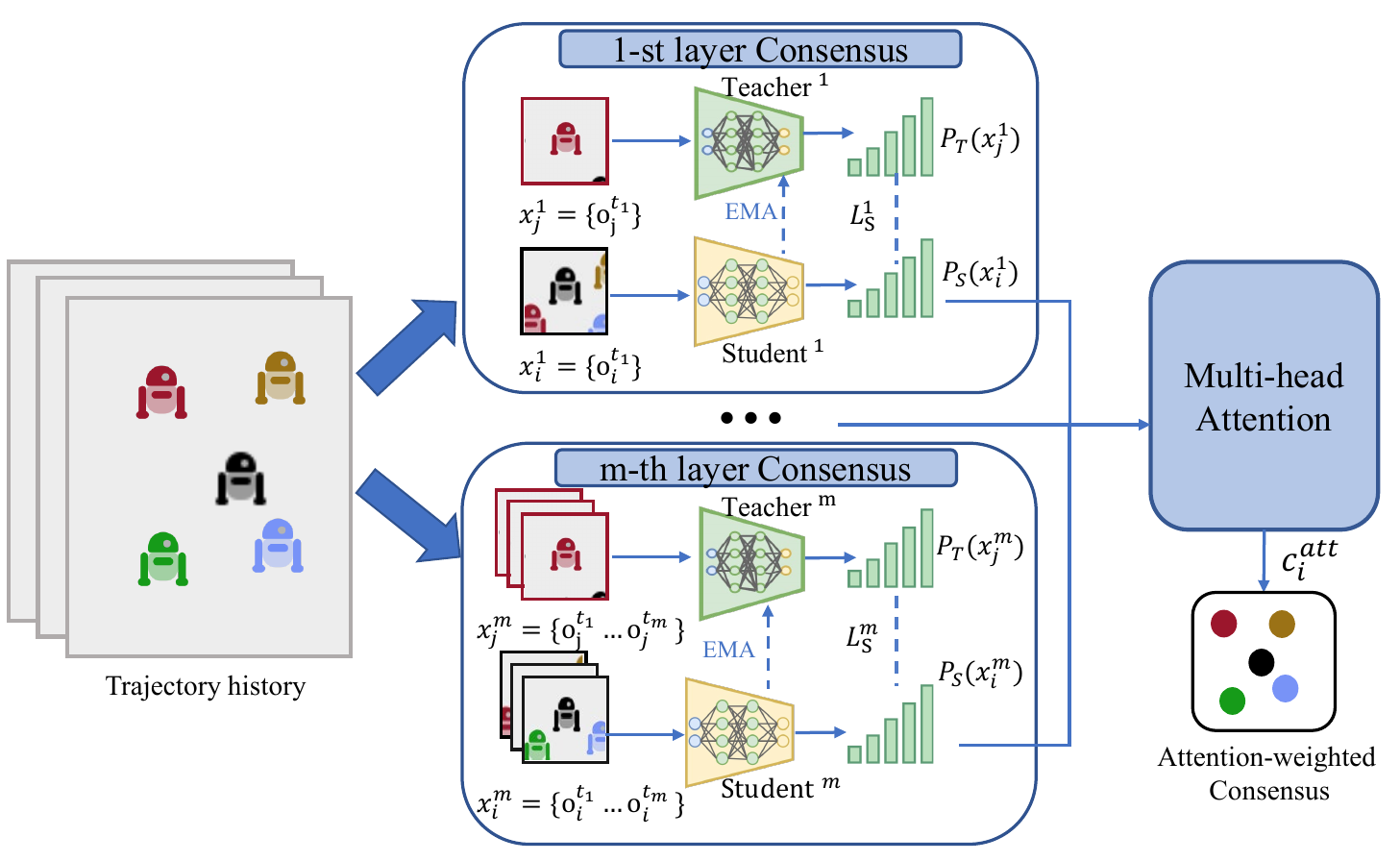}
\caption{An overview of the Hierarchical Consensus Mechanism. $x^m_i$ and $x^m_j$ represent different local observations from the same environmental state for the $m$-th layer, which are used to derive a global consensus classification through the teacher-student network. Consensus from different layers is aggregated into an attention-weighted consensus through multi-head attention.}
\label{fig:hierarchical_consensus_mechanism}
\label{Hierarchical Consensus Mechanism}
\vspace{-0.25in}
\end{figure}

\subsection{Hierarchical Consensus Mechanism}
Through the consensus builder, we have obtained a global consensus among agents based on their local observations. However, as shown in Fig.~\ref{hc}, obtaining a complete and effective global consensus from a single moment's local observations is challenging in practical applications. To address this issue, this section introduces a hierarchical consensus mechanism, divided into short-term consensus and long-term consensus. Short-term consensus considers only the current timestep's state, while long-term consensus takes into account information across multiple timesteps, incorporating a longer-term utilization of historical state information. This is dynamically leveraged through an attention mechanism that weighs the importance of short-term and long-term consensus. For instance, in scenarios requiring immediate collision avoidance, agents prioritize short-term consensus. Conversely, in collaborative search tasks, agents rely more on long-term consensus to allocate search areas efficiently.

We expand the foundational consensus builder into a Hierarchical Consensus Mechanism, distinguishing between short-term and long-term consensus. Short-term consensus leverages observations or consensus from a single timestep. For long-term consensus, we introduce \(x^m_i = \{o_i^{t_1}, o_i^{t_2}, o_i^{t_3}, \ldots, o_i^{t_m}\}\), a set representing the agent \(i\)'s observations at various timesteps within the trajectory history, where \(m\) indicates the inclusion of \(m\) historical observations. It's crucial to note that \(t_1\), \(t_2\), \(t_3\), and so forth, denote distinct timesteps, which are not required to be consecutive. This approach proves especially beneficial in scenarios with brief training intervals, where successive states may exhibit minimal differences. By considering observations at spaced intervals, we capture more pronounced changes over time, thereby gaining insights into significant state transitions.

As we transition from utilizing single-timestep observations to aggregating multi-timestep observations for contrastive learning, the optimization criterion for the \(m\)-th layer student network is correspondingly revised. The updated loss function is defined as follows:
\begin{equation}
L^m_{S}(\theta) = -\sum_{i=1}^{n}\sum_{j=1}^{n}\sum_{k} P_T(x^m_j)_k \log P_S(x^m_i)_k
\end{equation}
Therefore, the consensus for the \(m\)-th layer, \(c^m\), is redefined for each agent \(i\) as follows:
\begin{equation}
c^m_i = \arg\max_k P_S(x^m_i)_k
\end{equation}

Given the multiple layers of consensus achieved through the hierarchical structure, it becomes crucial to evaluate and weigh these layers differently across various scenarios. This differentiation acknowledges that the importance of short-term and long-term consensus can vary significantly depending on the context. To address this, we incorporate an attention mechanism that treats the consensus from each layer as input, as depicted in Fig.~\ref{Hierarchical Consensus Mechanism}. This approach enables us to integrate these varied consensus inputs into a multi-head attention framework, effectively allowing for the dynamic weighting of each layer's consensus.

To formalize this approach, we consider the output consensus from each layer, \(c^m_i\), as input to a multi-head attention mechanism. This mechanism aims to dynamically weigh the consensus from different layers according to the current scenario's specific requirements, resulting in a contextually weighted combination. The formalization is given by:

\begin{equation}
c^{att}_i = \text{MultiHead}(Q(c^m_i), K(c^m_i), V(c^m_i))
\end{equation}

In this equation, \(c^{att}_i\) represents the attention-weighted consensus for agent \(i\). The functions \(Q\), \(K\), and \(V\) correspond to the query, key, and value functions, respectively, which are applied to the consensus inputs. These functions facilitate the mapping of consensus from each layer into a space that allows for the evaluation of the layers' relevance. The MultiHead attention mechanism aggregates these mapped representations, allocating weights based on their assessed importance to the current decision-making context.

\begin{figure}[t]
\centering
\includegraphics[scale=0.4]{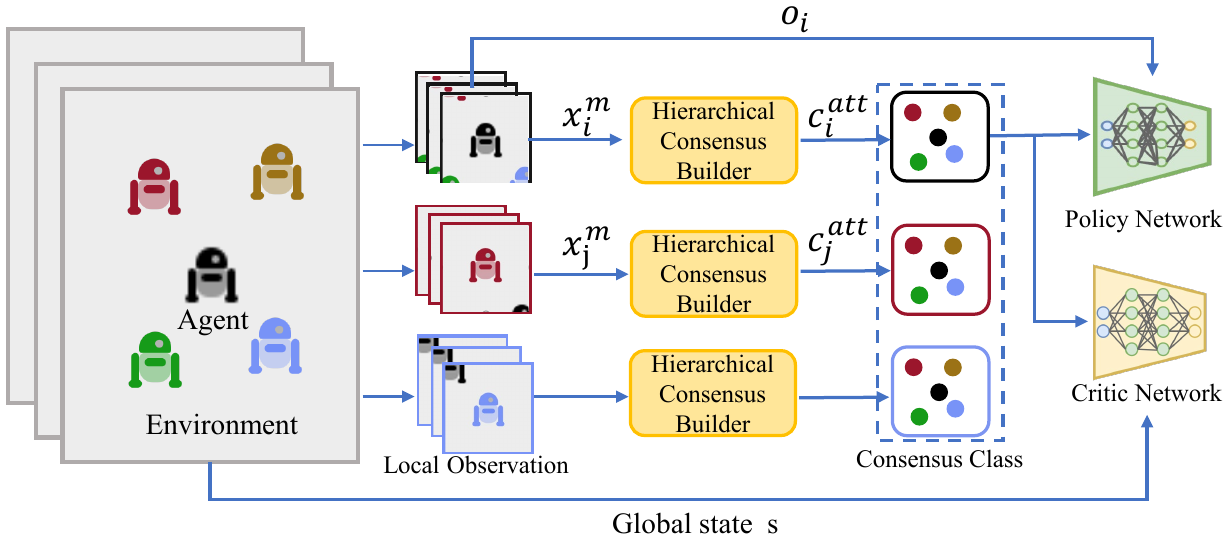}
\caption{Overview of the HC-MARL framework. Sequentially, from left to right: Agents initially acquire local observations from the environment. These observations are subsequently processed by the hierarchical consensus builder, yielding the current consensus class. This derived consensus, denoted as \(c^{att}_i\), enriches the agents' observational or state data. It is then incorporated into both policy and critic networks, thereby steering agent actions in alignment with the collectively determined global consensus.}
\label{framework}
\vspace{-0.2in}
\end{figure}


\subsection{HC-MARL Framework}

As illustrated in Fig.~\ref{framework}, the hierarchical consensus mechanism enables us to derive the attention-weighted consensus, \(c^{att}_i\). This consensus serves as the agent's inferred understanding of the global state, derived from partial observations. We integrate this consensus as an augmented observation input within the multi-agent reinforcement learning framework. It's pivotal to emphasize that while information from other agents is leveraged during the training phase of the student network, the action execution phase exclusively relies on an agent's local observations. This design principle ensures that our HC-MARL approach is compatible with a wide range of MARL algorithms, adhering to the CTDE paradigm.

\begin{figure*}[h]  
    \centering
    \begin{subfigure}{0.31\textwidth}
        \includegraphics[width=\linewidth]{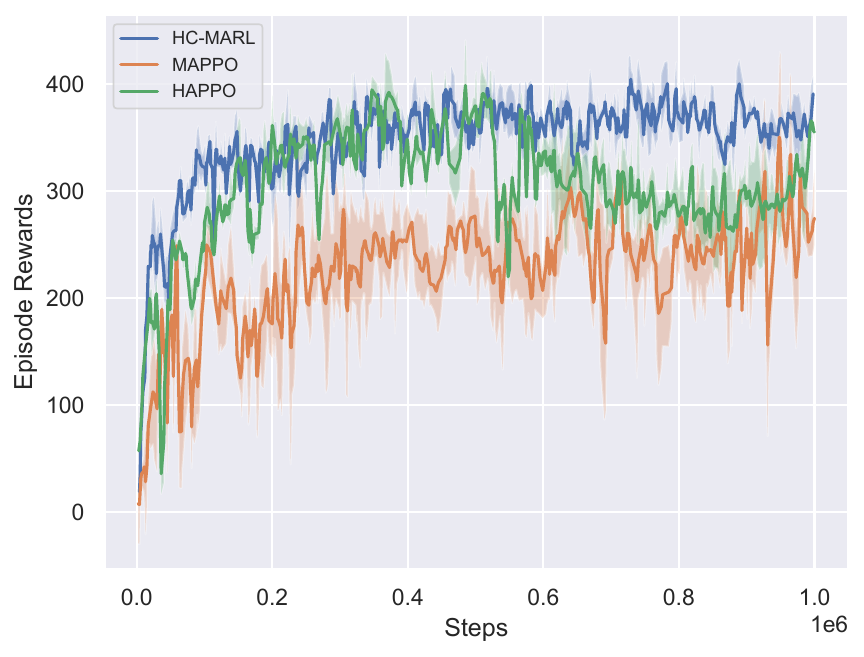}
        \caption{3 Predators - 1 Prey}
        \label{pp3}
    \end{subfigure}%
    \hfill  
    \begin{subfigure}{0.31\textwidth}
        \includegraphics[width=\linewidth]{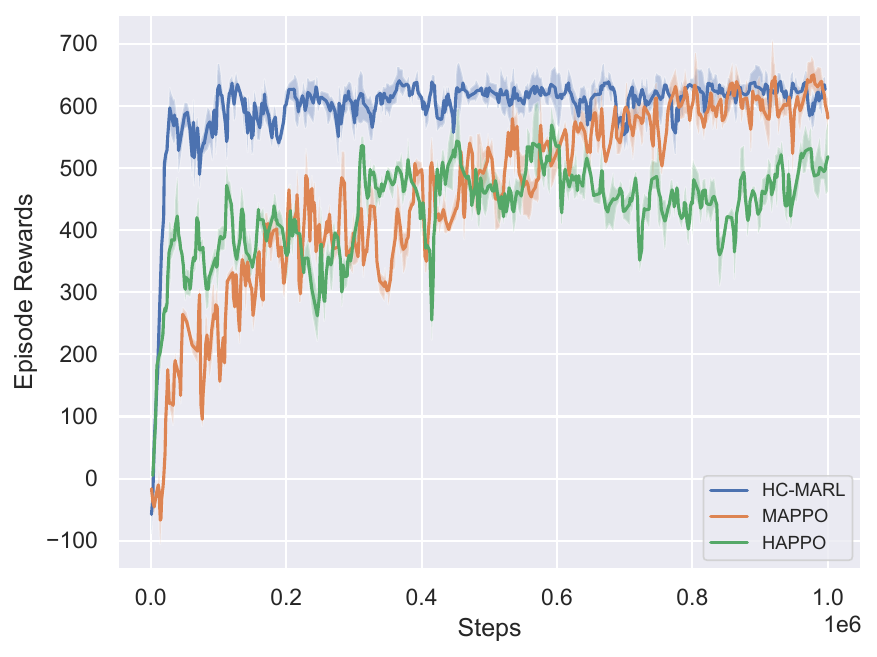}
        \caption{5 Predators - 1 Prey}
        \label{pp5}
    \end{subfigure}%
    \hfill
    \begin{subfigure}{0.31\textwidth}
        \includegraphics[width=\linewidth]{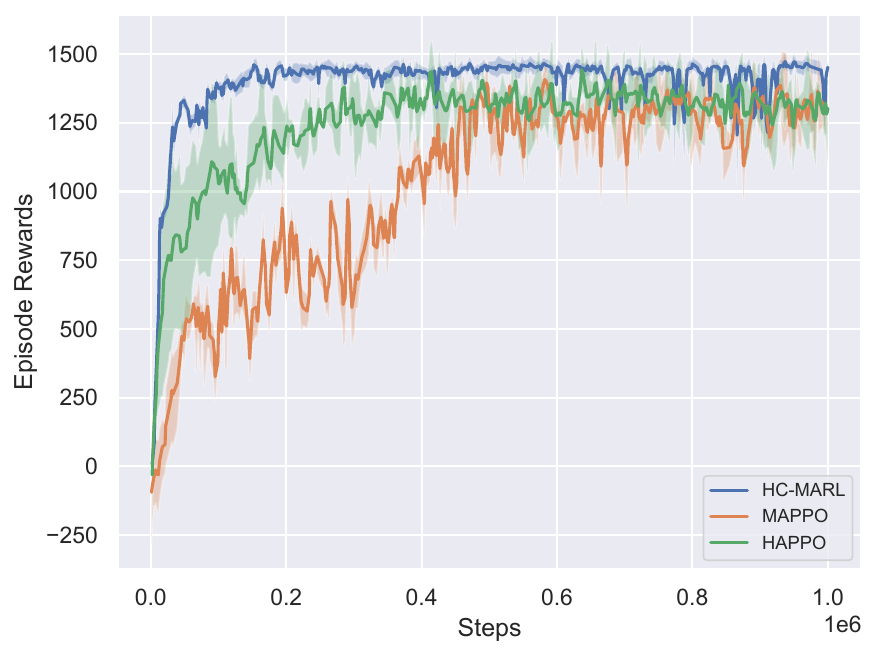}
        \caption{10 Predators - 1 Prey}
        \label{pp10}
    \end{subfigure}
    \caption{Learning curves of the HC-MARL, MAPPO, HAPPO on the Predator-Prey task. Each experiment was executed 5 times with different random seeds.}
    \label{pur reward curve}

\end{figure*}

\begin{figure*}[h]  
    \centering
    \begin{subfigure}{0.31\textwidth}
        \includegraphics[width=\linewidth]{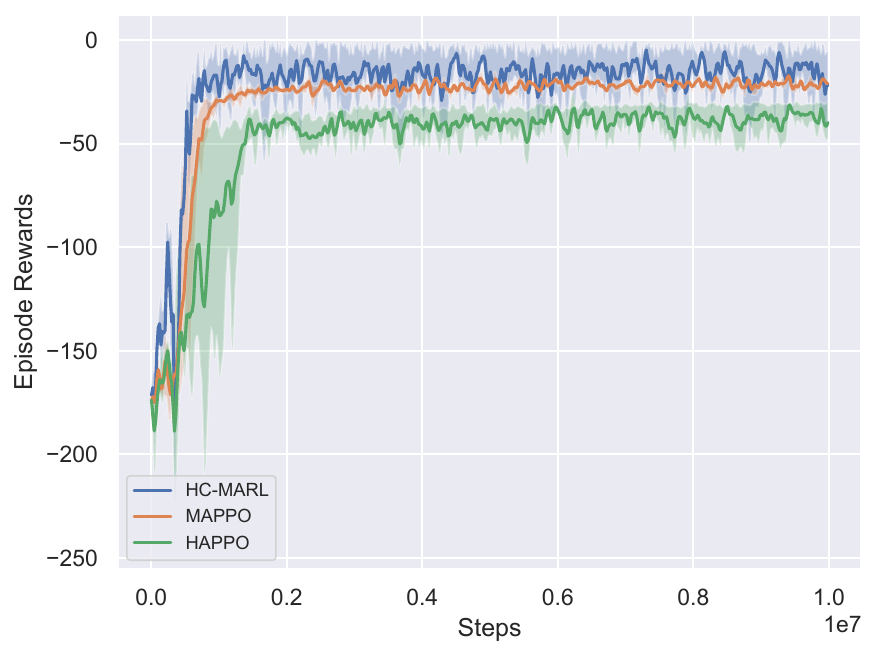}
        \caption{3 Agents}
    \end{subfigure}%
    \hfill  
    \begin{subfigure}{0.31\textwidth}
        \includegraphics[width=\linewidth]{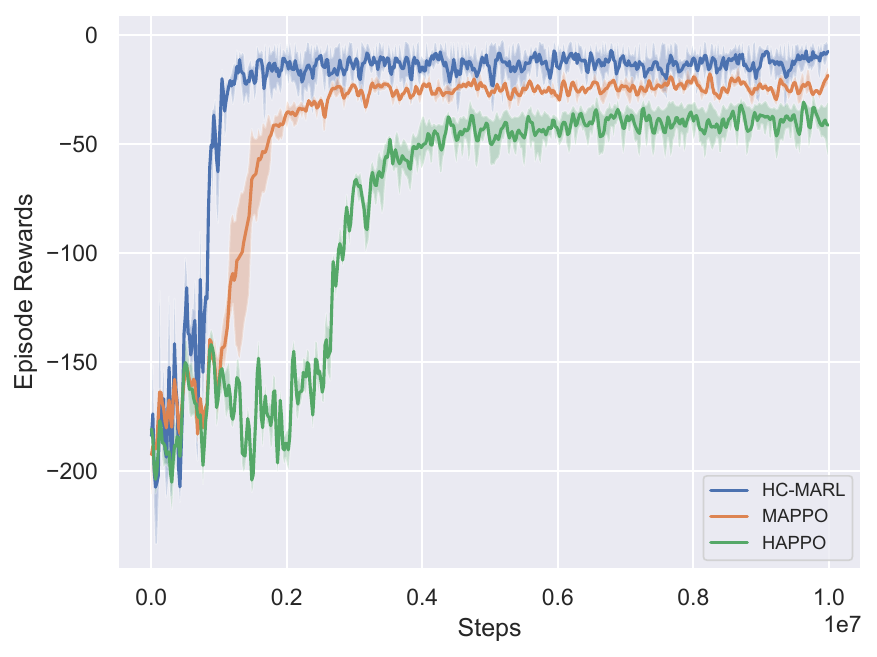}
        \caption{5 Agents}
    \end{subfigure}%
    \hfill
    \begin{subfigure}{0.31\textwidth}
        \includegraphics[width=\linewidth]{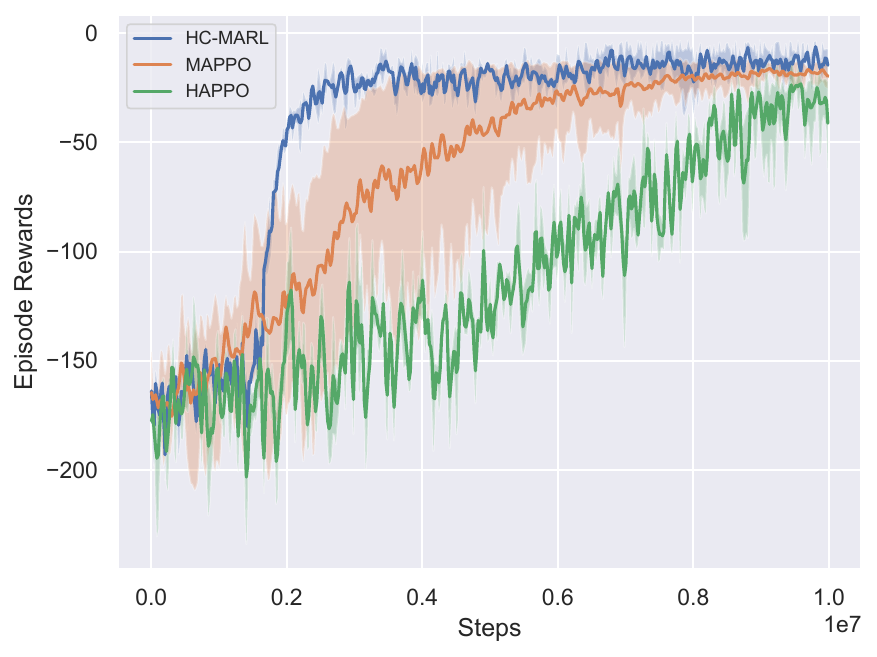}
        \caption{10 Agents}
    \end{subfigure}
    \caption{Learning curves of the HC-MARL, MAPPO, HAPPO on the Rendezvous task. Each experiment was executed 5 times with different random seeds.}
    \label{ren reward curve}

\end{figure*}

\begin{figure*}[h]  
    \centering
    \begin{subfigure}{0.31\textwidth}
        \includegraphics[width=\linewidth]{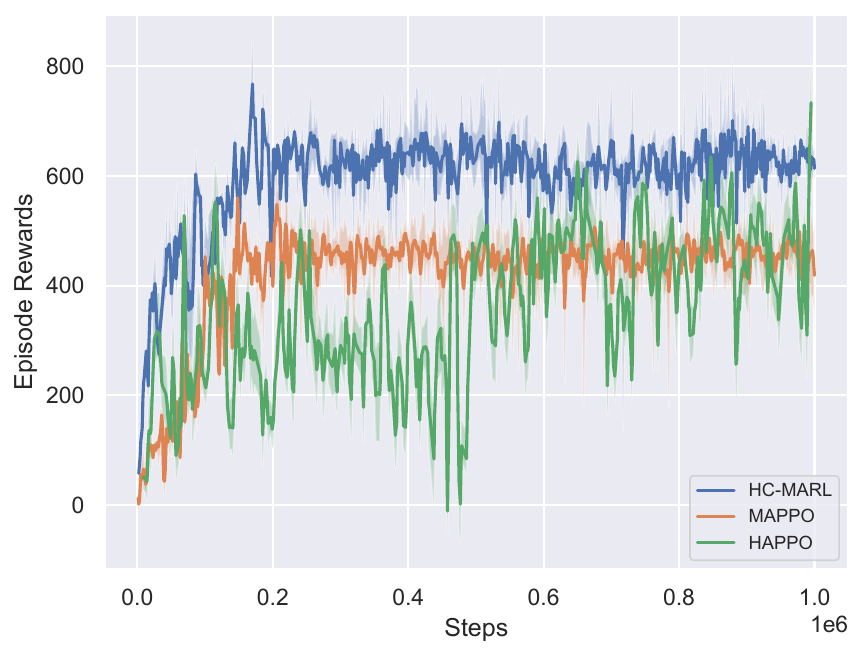}
        \caption{3 Agents}
    \end{subfigure}%
    \hfill  
    \begin{subfigure}{0.31\textwidth}
        \includegraphics[width=\linewidth]{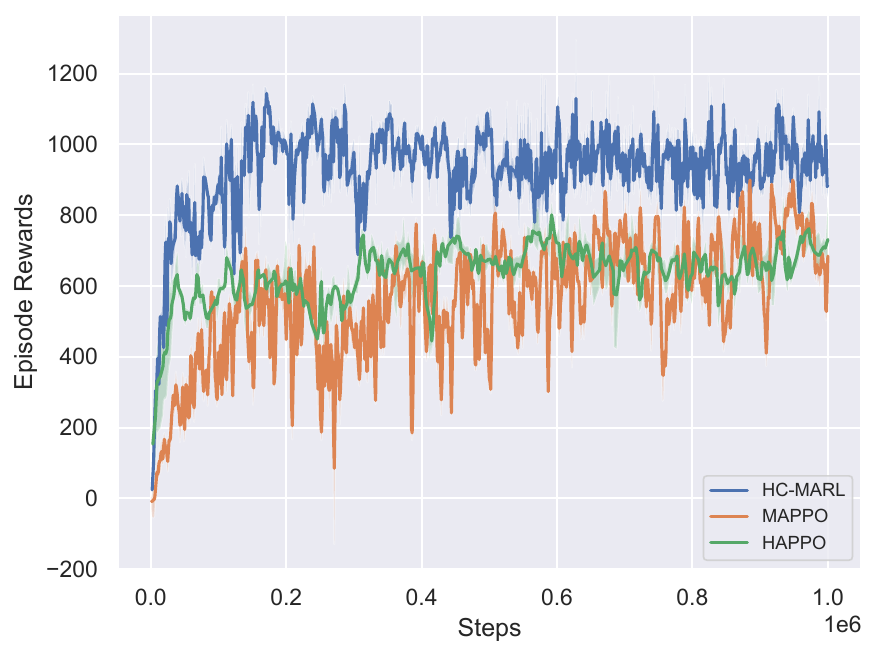}
        \caption{5 Agents}
    \end{subfigure}%
    \hfill
    \begin{subfigure}{0.31\textwidth}
        \includegraphics[width=\linewidth]{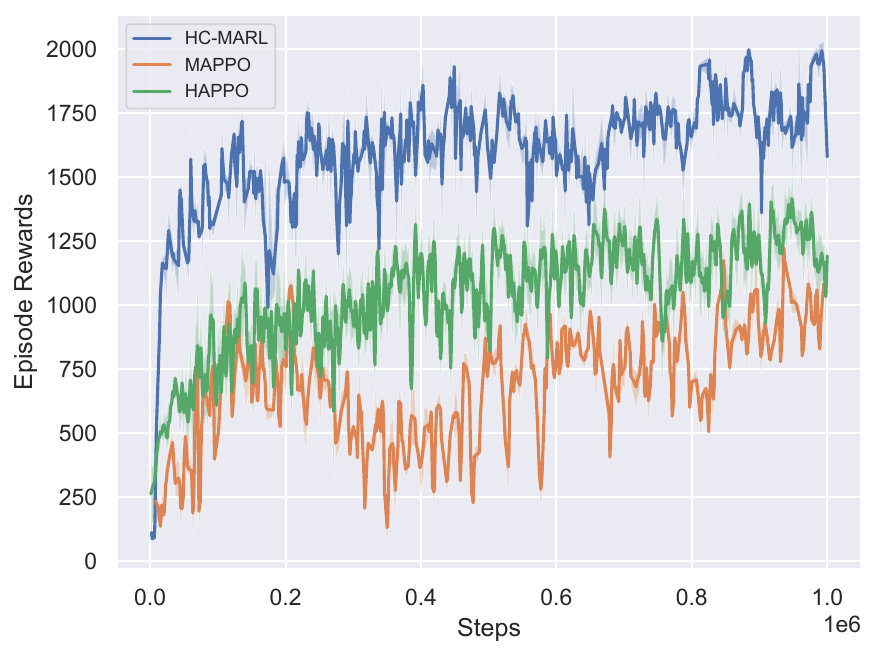}
        \caption{10 Agents}
    \end{subfigure}
    \caption{Learning curves of the HC-MARL, MAPPO, HAPPO on the Navigation task. Each experiment was executed 5 times with different random seeds.}
    \vspace{-0.25in}
    \label{navi reward curve}

\end{figure*}

Taking MAPPO as an example, by incorporating consensus information into the observations, the update for the critic network becomes:

\begin{align}
L^{\text{Critic}}(\phi) = &\ \mathbb{E}_{(s, \mathbf{a}, r, s')} \Big[ \Big( Q_\phi(s, c^{att}, \mathbf{a}) \nonumber \\
&- (r + \gamma Q_{\phi'}(s', c^{att'}, \mathbf{a}')) \Big)^2 \Big]
\end{align}

where \(c^{att}\) and \(c^{att'}\) represent the attention-weighted consensus information at the current and next time steps, respectively. The actor network update is formulated as follows:
\begin{equation}
\nabla_{\psi} J(\pi) = \mathbb{E}_{o,c^{att}, \mathbf{a} \sim \rho^\pi} \left[ \nabla_{\theta} \log \pi(o,c^{att}, \mathbf{a}|\psi) Q_\phi(s,c^{att}, \mathbf{a}) \right]
\end{equation}

These formulations illustrate how consensus information, specifically the attention-weighted consensus \(c^{att}\), is seamlessly integrated into the MARL process, enhancing the learning mechanism by providing a more informed perspective on the environment.

\section{Experiment and Results}
We conducted both simulations and hardware experiments to validate the efficacy of our proposed HC-MARL. 

\subsection{Environment Settings}

\begin{figure}[H]
  \centering
  \begin{subfigure}{0.32\linewidth}
    \centering
    \includegraphics[width=\linewidth]{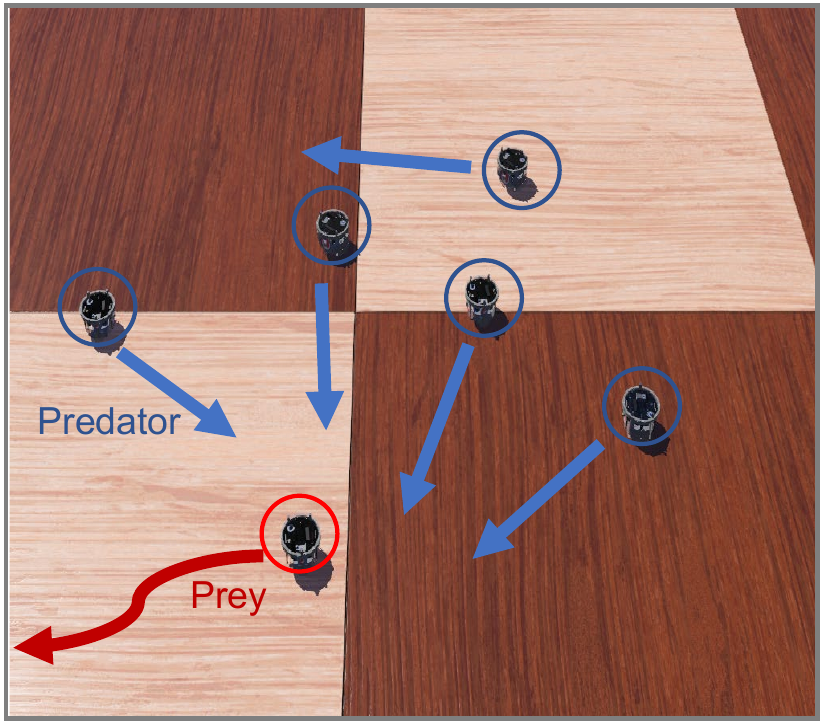}
    \caption{\small Predator-Prey}
    \label{tra1}
  \end{subfigure}
  \begin{subfigure}{0.32\linewidth}
    \centering
    \includegraphics[width=\linewidth]{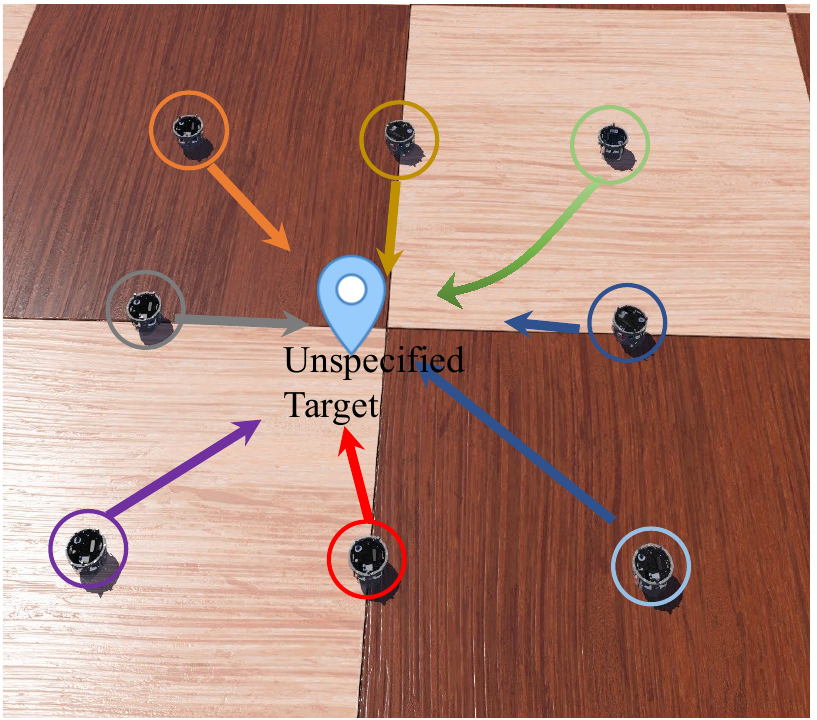}
    \caption{\small Rendezvous}
    \label{tra3}
  \end{subfigure}
  \begin{subfigure}{0.32\linewidth}
    \centering
    \includegraphics[width=\linewidth]{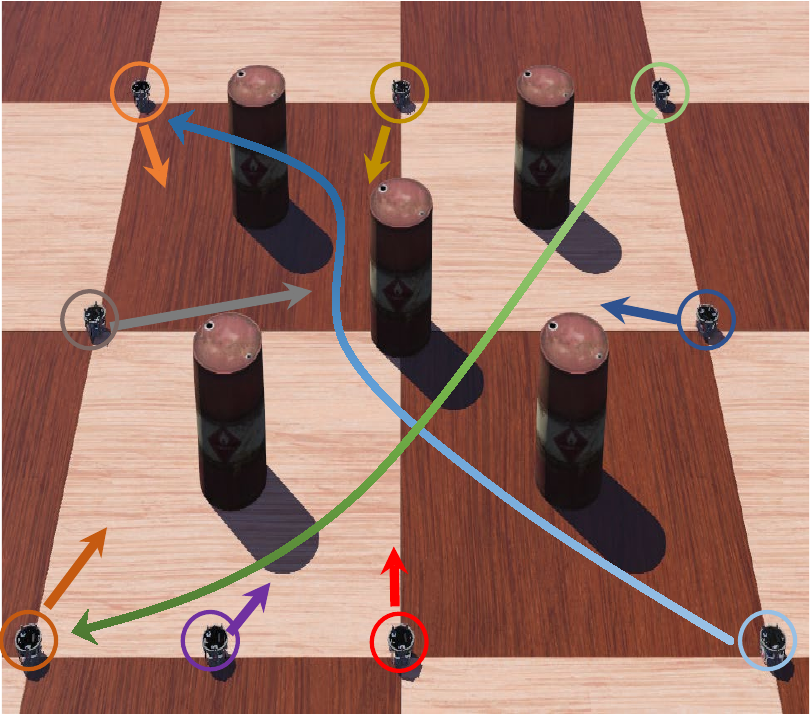}
    \caption{\small Navigation}
    \label{tra4}
  \end{subfigure}
  \caption{The simulated tasks considered in the experiments.} 
  \vspace{-0.15in}
  \label{3_task} 
\end{figure}
We constructed three cooperative multi-agent tasks within the Webots simulation, including Predator-Prey, Rendezvous, and Navigation, as shown in Fig.~\ref{3_task}. We implemented our HC-MARL in these three tasks and compared it against two main-stream MARL baselines Multi-Agent Proximal Policy Optimization (MAPPO) and its variant Heterogeneous-Agent Proximal Policy Optimization (HAPPO)~\cite{zhong2024heterogeneous}. Note that our HC-MARL framework can seamlessly integrate with various MARL algorithms. To ensure fair comparisons with these variants of MAPPO, we constructed our HC-MARL framework based on the MAPPO architecture in this experiment.

\subsection{Main Results}

This section describes and analyzes the experimental results in three tasks. The performance of each algorithm was evaluated with five different random seeds.  The learning curves, in terms of episode reward under varying numbers of agents, are presented in Fig.~\ref{pur reward curve}, \ref{ren reward curve}, and \ref{navi reward curve}. In addition to episode rewards, Table~\ref{performance_table} compares the differences in algorithm performance through the number of steps required to complete the tasks after training. These results demonstrate that our work, HC-MARL, achieved varying degrees of advantage over all baseline algorithms.


\textbf{Predator-Prey task.} In this scenario, predators must pursue and catch the prey through movement. The number of predators was set to 3, 5, and 10, respectively, with the number of prey fixed at one. The prey's escape trajectory was randomly generated during both the training and testing phases. Fig.~\ref{pp3},~\ref{pp5}, and~\ref{pp10} respectively showcase the learning curves under the settings of three, five, and ten predators. The results demonstrate that our work, HC-MARL, surpasses the baseline algorithms in terms of episode reward convergence and convergence speed. Furthermore, Table~\ref{performance_table} reveals that HC-MARL requires significantly fewer steps to complete the task post-training compared to the baseline algorithms, indicating that our method accomplishes tasks more efficiently and enhances the performance of the algorithm.



\textbf{Rendezvous task. } In the Rendezvous task, where no specific target points are assigned, agents autonomously aggregate from their random initial locations on the map. The number of agents was set to 3, 5, and 10. From Fig.~\ref{ren reward curve}, it can be observed that HC-MARL outperforms the baselines in terms of convergence episode rewards and convergence speed. The performance advantage of HC-MARL over the baselines becomes more pronounced with an increasing number of agents. Analyzing this phenomenon, it is evident that as the number of agents increases, and thus the complexity of the task escalates, the hierarchical consensus mechanism contributes more significantly to enhancing the algorithm's performance. Table~\ref{performance_table} also demonstrates that HC-MARL surpasses the baseline algorithms in terms of the number of steps required to complete the task.



\begin{table*}[ht]
\vspace{0.2cm}
\caption{Number of steps required to complete the Task for HC-MARL and baselines (after training) across three tasks. Error bars indicate the standard error of the mean.}
\label{performance_table}
\centering
\resizebox{1.0\textwidth}{!}{
    \begin{tabular}{c|ccc|ccc|ccc}
    \hline
\toprule
 Steps&\multicolumn{3}{c|}{Predator-Prey}  & \multicolumn{3}{c|}{Rendezvous} & \multicolumn{3}{c}{Navigation}\\ \hline
           Agents&  MAPPO&  HAPPO&  HC-MARL(ours) &  MAPPO&  HAPPO&  HC-MARL(ours) &MAPPO&  HAPPO&  HC-MARL(ours) \\\hline
           3 &  $720 \pm 60$&  $740 \pm 50$&  \textbf{580} $\pm$ \textbf{45}&  $ 575 \pm 25$&  $ 585 \pm 35$&  \textbf{550} $\pm$ \textbf{25}&  $635 \pm 55$& $630 \pm 70$ & \textbf{520} $\pm$ \textbf{40}\\
           5 &  $550 \pm 65$&  $640 \pm 60$&  \textbf{510} $\pm$ \textbf{55}&  $640 \pm 35$&  $645 \pm 40$&   \textbf{610} $\pm$ \textbf{40}&  $710 \pm 60$& $680\pm 70$ & \textbf{590} $\pm$ \textbf{65}\\
           10&  $520\pm 60$&  $530\pm 60$&  \textbf{450}$\pm$ \textbf{55}&  $670\pm 35$&  $695\pm 45$&  \textbf{620} $\pm$ \textbf{45}&  $960\pm 60$& $890\pm 75$ &\textbf{700} $\pm$ \textbf{65} \\
    \bottomrule
\end{tabular}}
\vspace{-0.15in}
\end{table*}

\textbf{Navigation task. } In the navigation task, agents are asked to navigate through two obstacles and reach the target point while avoiding collisions with other agents and obstacles. Fig.~\ref{navi reward curve} demonstrates that our HC-MARL method significantly improves episode rewards across tasks with varying numbers of agents. Specifically, HC-MARL's episode rewards are approximately 20\% higher than those of HAPO and MAPPO in tasks with three agents, and about 35\% higher in tasks with ten agents. Furthermore, Table~\ref{performance_table} indicates that at ten agents, the improvement in the number of steps required to complete the obstacle navigation task is even more substantial. HC-MARL requires only 700 steps to complete the task, representing a reduction of 30\% and 40\% compared to HAPO and MAPPO, respectively.

\subsection{Ablation Study}


To assess the efficacy of both the consensus mechanism and the hierarchical approach, we conducted ablation studies varying the number of consensus categories and layers.

\begin{figure}[h]
  \centering
  \begin{subfigure}{0.48\linewidth}
    \centering
    \includegraphics[width=\linewidth]{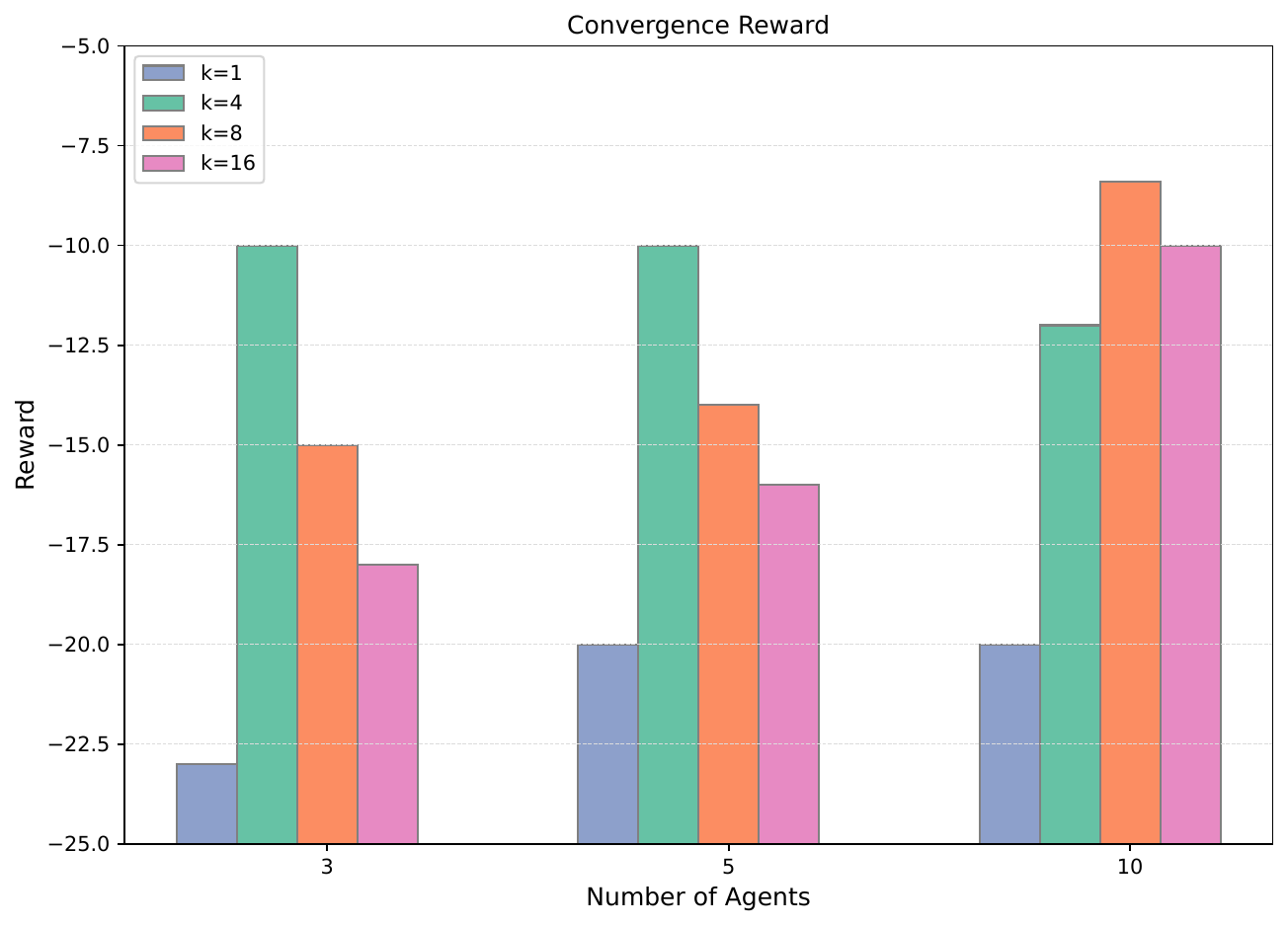}
    \caption{\small Influence of $k$}
    \label{k}
  \end{subfigure}
  \begin{subfigure}{0.48\linewidth}
    \centering
    \includegraphics[width=\linewidth]{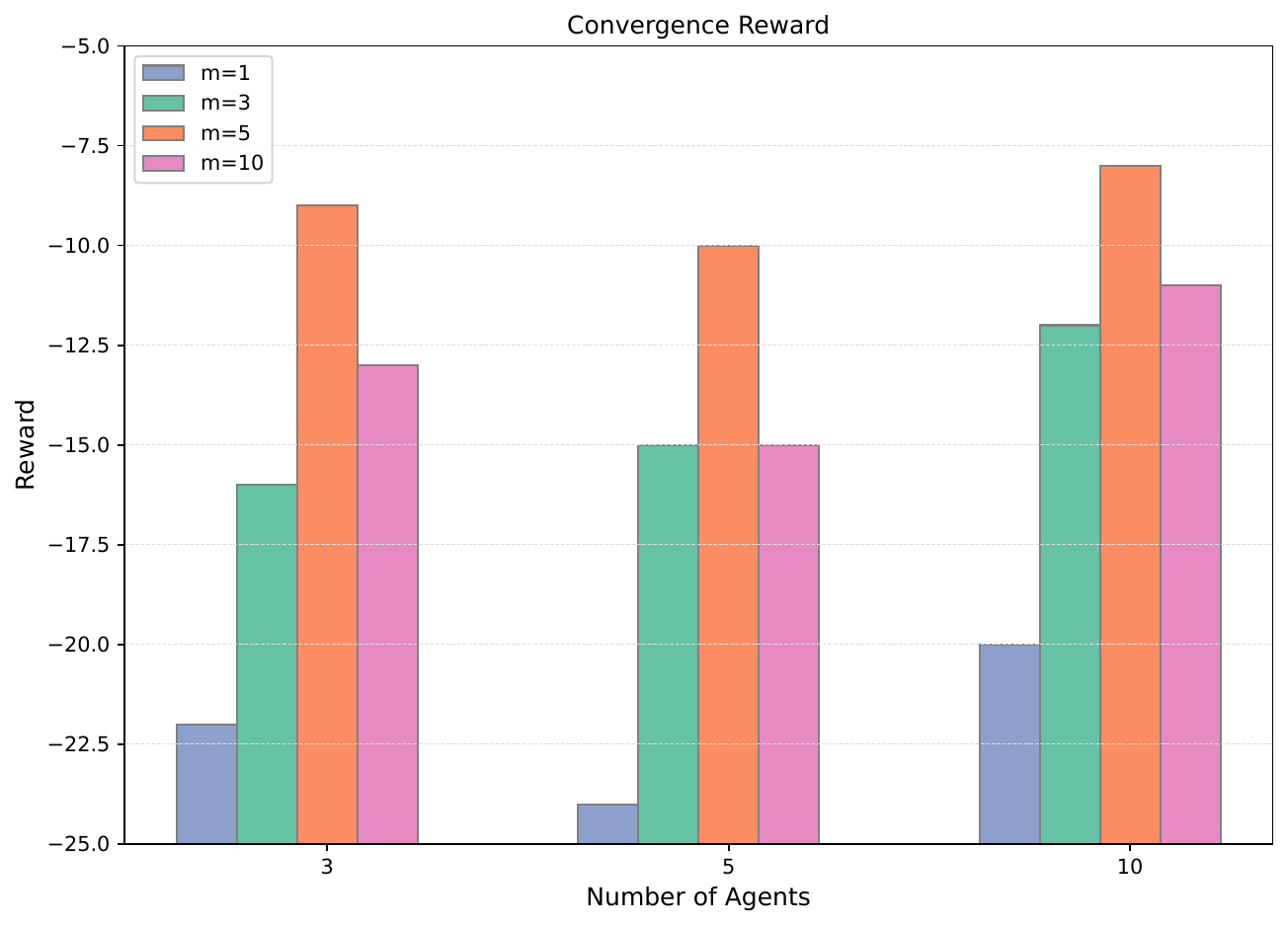}
    \caption{\small Influence of $m$}
    \label{m}
  \end{subfigure}
  \caption{Ablation study on HC-MARL in the Rendezvous task} 
  \vspace{-0.2in}
\end{figure}

Initially, we investigated the effect of global consensus categories \(k\) on the Rendezvous task, testing \(k\) values of 1, 4, 8, and 16 across agent counts of 3, 5, and 10. Fig.~\ref{k} shows scenarios with \(k>1\) yield higher convergence rewards than those with \(k=1\), highlighting the consensus mechanism's benefit. Optimal rewards for 3 and 5 agents occurred at \(k=4\); for 10 agents, \(k=8\) was most effective. This indicates that simpler scenarios benefit from fewer categories, while more agents necessitate more categories for optimal training.

Additionally, the impact of consensus layers \(m\) on training rewards was examined for \(m=1\) (no hierarchy), 3, 5, and 10, depicted in Fig.~\ref{m}. Optimal rewards were achieved at \(m=5\), suggesting that increasing consensus layers up to a point enhances task performance. However, beyond this optimal level, training efficiency declined, due to increased training complexity and instability with additional layers.

\subsection{Real World Experiments}
We validated the real-world applicability of HC-MARL by conducting experiments on E-puck swarm. We utilized the NOKOV motion capture system for indoor positioning. 

\begin{figure}[h]
\centering

\begin{subfigure}{0.48\columnwidth}
\includegraphics[width=\linewidth]{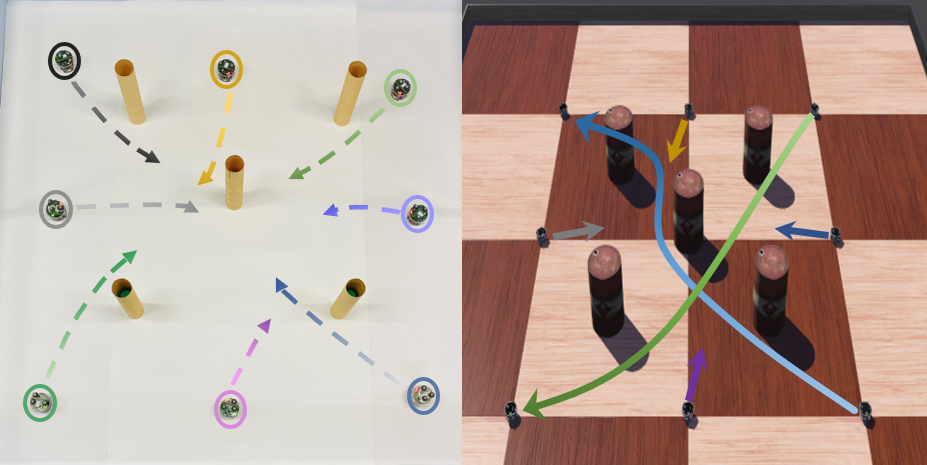}
\caption{Start}
\label{fig:image1}
\end{subfigure}
\hfill
\begin{subfigure}{0.48\columnwidth}
\includegraphics[width=\linewidth]{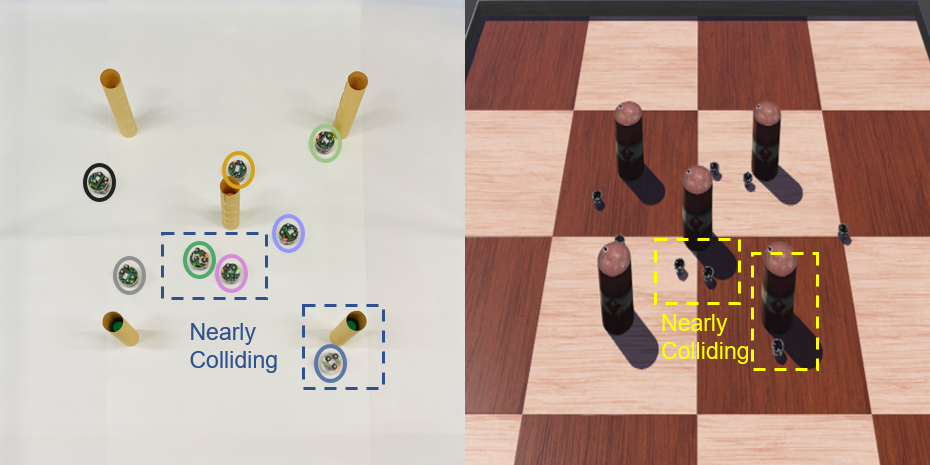}
\caption{Nearly Colliding}
\label{fig:image2}
\end{subfigure}

\vspace{1em} 

\begin{subfigure}{0.48\columnwidth}
\includegraphics[width=\linewidth]{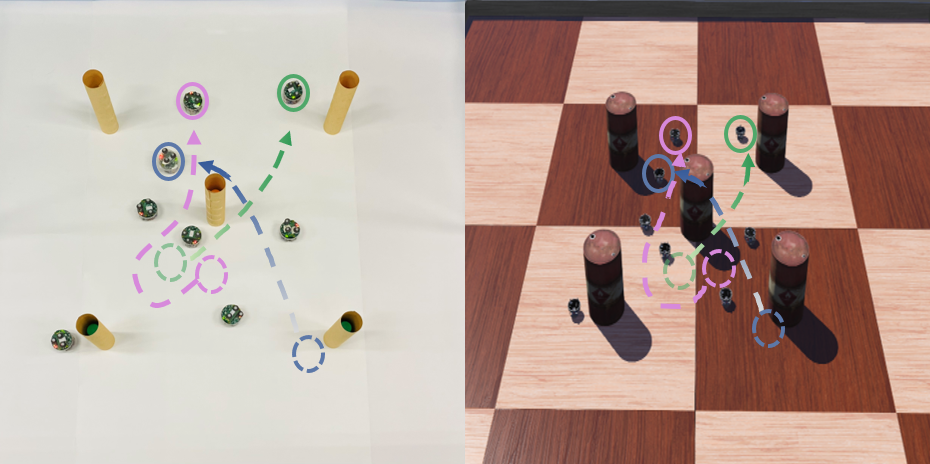}
\caption{Collision avoidance
}
\label{fig:image3}
\end{subfigure}
\hfill
\begin{subfigure}{0.48\columnwidth}
\includegraphics[width=\linewidth]{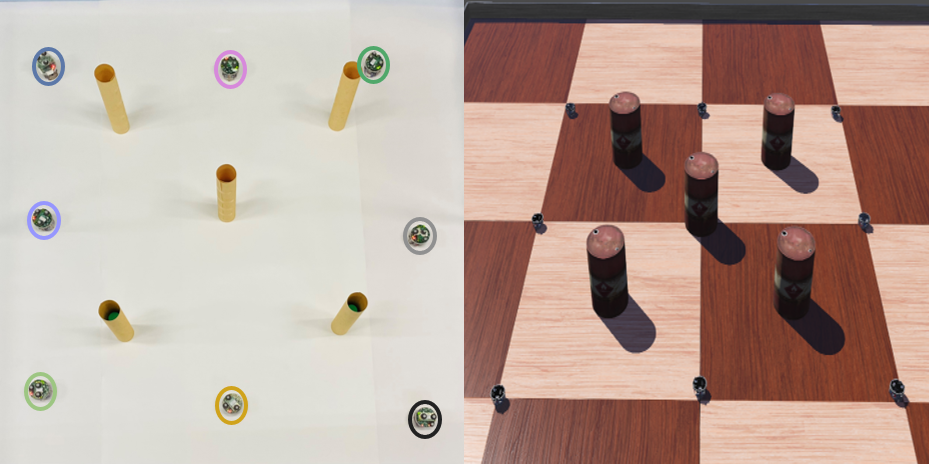}
\caption{Arrived}
\label{Arrived}
\end{subfigure}

\caption{Navigation Task Demonstrations. Left is the real-world environment, and Right is the Webots simulation.}
\label{real}
\vspace{-0.25in}
\end{figure}

We conducted experiments on Predator-Prey, Rendezvous, and Navigation tasks. In the Predator-Prey scenario, the HC-MARL algorithm required 16\% fewer steps to capture the prey than MAPPO and 19\% fewer than HAPPO. For the Rendezvous task, agents using HC-MARL completed the gathering objective with 10\% fewer steps compared to MAPPO and 15\% fewer than HAPPO. In the Navigation task, HC-MARL agents reached their target destinations with 30\% less distance traveled than MAPPO and 34\% less than HAPPO, without any collisions with obstacles. Taking the Navigation task as an example, Fig.~\ref{real} displays four representative scenes captured during both simulated and real-world experiments. For a visual representation, supplementary videos can be found in the Appendix.

\section{Conclusion}


We introduced the Hierarchical Consensus-Based Multi-Agent Reinforcement Learning (HC-MARL) framework, a novel approach that employs hierarchical consensus to facilitate cooperative execution among agents based on local observations. Recognizing that each agent's local observations are subsets of a consistent global state, our framework leverages contrastive learning from these observations to achieve a global consensus, which then serves as additional local observations for the agents. By implementing a hierarchical mechanism, we construct short-term and long-term consensus to cater to the dynamic requirements of various tasks. Extensive experiments demonstrate that the HC-MARL method significantly enhances the performance of multi-robot cooperation tasks.

\section{ACKNOWLEDGMENT}
We gratefully acknowledge the support of the National Science and Technology Major Project(No. 2022ZD0116401), the National Natural Science Foundation of China (Grant No. 62306023), and the Science and Technology Project of State Grid Corporation of China (No. 5108-202218280A-2-402-XG).





\bibliographystyle{IEEEtran}
\bibliography{ref}

\end{document}